# Supplementary Results for Named Entity Recognition on Chinese Social Media with an Updated Dataset

**Nanyun Peng** and **Mark Dredze**
Human Language Technology Center of Excellence
Center for Language and Speech Processing
Johns Hopkins University, Baltimore, MD, 21218
npeng1@jhu.edu, mdredze@cs.jhu.edu

Our paper Peng and Dredze (2015) introduced the task of Named Entity Recognition (NER) on Chinese Social Media with an accompanying dataset. We further improved the NER system by multi-task learning with Chinese Word Segmentation (Peng and Dredze, 2016).

Our NER annotations on Weibo messages was constructed using Amazon Mechanical Turk[1], and the final annotations were generated by merging labels from multiple different Turkers using heuristics. This inevitably lead to inconsistencies and errors in the dataset.

He and Sun (2017a) manually corrected the annotations which resulted in a much cleaner dataset. This corrected dataset is available on Github along with code from our original paper.[2]

This paper provides updated results for this corrected dataset using the method described in Peng and Dredze (2015) and Peng and Dredze (2016). This will allow future work to compare to our results and use the cleaner dataset.

We tuned the hyper-parameters as discussed in Peng and Dredze (2015) and Peng and Dredze (2016). We only show the best results obtained from the best model proposed in Peng and Dredze (2015) and Peng and Dredze (2016). Table 1 shows that we get much better results and outperformed both He and Sun (2017a) and He and Sun (2017b).

To facilitate comparison with He and Sun (2017a), we report results using their format.

| Models | Named Entity | | | Nominal Mention | | | Overall |
|---|---|---|---|---|---|---|---|
| | Prec | Recall | F1 | Prec | Recall | F1 | |
| He and Sun (2017a) | 66.93 | 40.67 | 50.60 | 66.46 | 53.57 | 59.32 | 54.82 |
| He and Sun (2017b) | 61.68 | 48.82 | 54.50 | 74.13 | 53.54 | 62.17 | 58.23 |
| Peng and Dredze (2015) | 74.78 | 39.81 | 51.96 | 71.92 | 53.03 | 61.05 | 56.05 |
| Peng and Dredze (2016) | 66.67 | 47.22 | 55.28 | 74.48 | 54.55 | 62.97 | 58.99 |

Table 1: Test results for Peng and Dredze (2015) and Peng and Dredze (2016) on the updated Chinese Social Media NER dataset. We got much better results than the originally reported number. We also listed the results in He and Sun (2017a) and He and Sun (2017b) for comparison purposes.

## Acknowledgements

We thank Hangfeng He and Xu Sun for correcting our dataset and sharing the new annotations.

---

[1] https://www.mturk.com/mturk/welcome
[2] https://github.com/hltcoe/golden-horse

# Improving Named Entity Recognition for Chinese Social Media with Word Segmentation Representation Learning

**Nanyun Peng**\* and **Mark Dredze**\*†
\* Human Language Technology Center of Excellence
Center for Language and Speech Processing
Johns Hopkins University, Baltimore, MD, 21218
†Bloomberg LP, New York, NY 10022
npeng1@jhu.edu, mdredze@cs.jhu.edu

## Abstract

Named entity recognition, and other information extraction tasks, frequently use linguistic features such as part of speech tags or chunkings. For languages where word boundaries are not readily identified in text, word segmentation is a key first step to generating features for an NER system. While using word boundary tags as features are helpful, the signals that aid in identifying these boundaries may provide richer information for an NER system. New state-of-the-art word segmentation systems use neural models to learn representations for predicting word boundaries. We show that these same representations, jointly trained with an NER system, yield significant improvements in NER for Chinese social media. In our experiments, jointly training NER and word segmentation with an LSTM-CRF model yields nearly 5% absolute improvement over previously published results.

## 1 Introduction

Entity mention detection, and more specifically named entity recognition (NER) (Collins and Singer, 1999; McCallum and Li, 2003; Nadeau and Sekine, 2007; Jin and Chen, 2008; He et al., 2012), has become a popular task for social media analysis (Finin et al., 2010; Liu et al., 2011; Ritter et al., 2011; Fromreide et al., 2014; Li et al., 2012; Liu et al., 2012a). Many downstream applications that use social media, such as relation extraction (Bunescu and Mooney, 2005) and entity linking (Dredze et al., 2010; Ratinov et al., 2011), rely on first identifying mentions of entities. Not surprisingly, accuracy of NER systems in social media trails state-of-the-art systems for news text and other formal domains. While this gap is shrinking in English (Ritter et al., 2011; Cherry and Guo, 2015), it remains large in other languages, such as Chinese (Peng and Dredze, 2015; Fu et al., 2015).

One reason for this gap is the lack of robust up-stream NLP systems that provide useful features for NER, such as part-of-speech tagging or chunking. Ritter et al. (2011) annotated Twitter data for these systems to improve a Twitter NER tagger, however, these systems do not exist for social media in most languages. Another approach has been that of Cherry and Guo (2015) and Peng and Dredze (2015), who relied on training unsupervised lexical embeddings in place of these up-stream systems and achieved state-of-the-art results for English and Chinese social media, respectively. The same approach was also found helpful for NER in the news domain (Collobert and Weston, 2008; Turian et al., 2010; Passos et al., 2014)

In Asian languages like Chinese, Japanese and Korean, word segmentation is a critical first step for many tasks (Gao et al., 2005; Zhang et al., 2006; Mao et al., 2008). Peng and Dredze (2015) showed the value of word segmentation to Chinese NER in social media by using character positional embeddings, which encoded word segmentation information.

In this paper, we investigate better ways to incorporate word boundary information into an NER system for Chinese social media. We combine the state-of-the-art Chinese word segmentation system (Chen et al., 2015) with the best Chinese social media NER model (Peng and Dredze, 2015). Since both systems used learned representations, we propose an integrated model that allows for joint training learned representations, providing more information to the NER system about hidden representations learned from word segmentation, as compared to features based on segmentation output. Our integrated model achieves nearly a 5% absolute improvement over the previous best results on both NER and nominal mentions for Chinese social media.

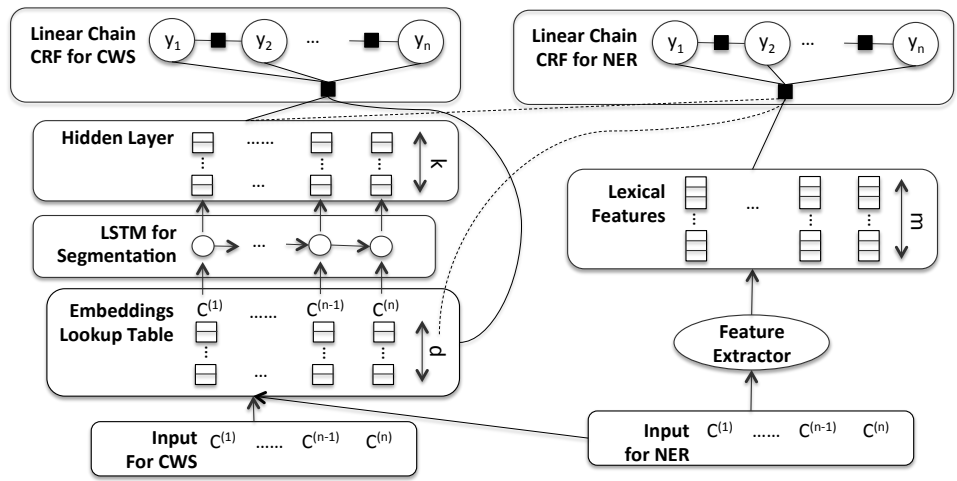

Figure 1: The joint model for Chinese word segmentation and NER. The left hand side is an LSTM module for word segmentation, and the right hand side is a traditional feature-based CRF model for NER. Note that the linear chain CRF for NER has both access to the feature extractor specifically for NER and the representations produced by the LSTM module for word segmentation. The CRF in this version is a log-bilinear CRF, where it treats the embeddings and hidden vectors inputs as variables and modifies them according to the objective function. As a result, it enables propagating the gradients back into the LSTM to adjust the parameters. Therefore, the word segmentation and NER training share all the parameters of the LSTM module. This facilitates the joint training.

## 2 Model

We propose a model that integrates the best Chinese word segmentation system (Chen et al., 2015) using an LSTM neural model that learns representations, with the best NER model for Chinese social media (Peng and Dredze, 2015), that supports training neural representations by a log-bilinear CRF. We begin with a brief review of each system.

### 2.1 LSTM for Word Segmentation

Chen et al. (2015) proposed a single layer, left to right LSTM for Chinese word segmentation. An LSTM is a recurrent neural network (RNN) which uses a series of gates (input, forget and output gate) to control how memory is propagated in the hidden states of the model. For the Chinese word segmentation task, each Chinese character is initialized as a $d$ dimensional vector, which the LSTM will modify during its training. For each input character, the model learns a hidden vector $h$. These vectors are then used with a biased-linear transformation to predict the output labels, which in this case are **B**egin, **I**nside, **E**nd, and **S**ingleton. A prediction for position $t$ is given as:

$$y^{(t)} = W_o h^{(t)} + b_o \qquad (1)$$

where $W_o$ is a matrix for the transformation parameters, $b_o$ is a vector for the bias parameters, and $h^{(t)}$ is the hidden vector at position $t$. To model the tag dependencies, they introduced the transition score $A_{ij}$ to measure the probability of jumping from tag $i \in T$ to tag $j \in T$.

We used the same model as Chen et al. (2015) trained on the same data (segmented Chinese news article). However, we employed a different training objective. Chen et al. (2015) employed a max-margin objective, however, while they found this objective yielded better results, we observed that maximum-likelihood yielded better segmentation results in our experiments[1]. Additionally, we sought to integrate their model with a log-bilinear CRF, which uses a maximum-likelihood training objective. For consistency, we trained the LSTM with a maximum-likelihood training objective as well. The maximum-likelihood CRF objective function for predicting segmentations is:

---

[1]Chen et al. (2015) preprocessed the data specifically for Chinese word segmentation, such as replacing English characters, symbols, dates and Chinese idioms as special symbols. Our implementation discarded all these preprocessing steps, which while it achieved nearly identical results on development data (as inferred from their published figure), it lagged in test accuracy by 2.4%. However, we found that while these preprocessing steps improved segmentation, they hurt NER results as they resulted in a mis-match between the segmentation and NER input data. Since our focus is on improving NER, we do not use their preprocessing steps in this paper.

$$\mathcal{L}_s(\boldsymbol{y}_s; \boldsymbol{x}_s, \Theta) = \frac{1}{K} \sum_k \Big[ \log \frac{1}{Z(\boldsymbol{x}_s)^k} + \sum_i \Big( T_s(y_{i-1}^k, y_i^k) + s(y_i^k; \boldsymbol{x}_s^k, \boldsymbol{\Lambda}_s) \Big) \Big] \quad (2)$$

Example pairs $(\boldsymbol{y}_s, \boldsymbol{x}_s)$ are word segmented sentences, $k$ indexes examples, and $i$ indexes positions in examples. $T_s(y_{i-1}^k, y_i^k)$ are standard transition probabilities learned by the CRF[2]. The LSTM parameters $\Lambda_s$ are used to produce $s(y_i^k; \boldsymbol{x}_s^k, \Lambda_s)$, the emission probability of the label at position $i$ for input sentence $k$, which is obtained by taking a soft-max over (1). We use a first-order Markov model.

## 2.2 Log-bilinear CRF for NER

Peng and Dredze (2015) proposed a log-bilinear model for Chinese social media NER. They used standard NER features along with additional features based on lexical embeddings. By fine-tuning these embeddings, and jointly training them with a word2vec (Mikolov et al., 2013) objective, the resulting model is log-bilinear.

Typical lexical embeddings provide a single embedding vector for each word type. However, Chinese text is not word segmented, making the mapping between input to embedding vector unclear. Peng and Dredze (2015) explored several types of representations for Chinese, including pre-segmenting the input to obtain words, using character embeddings, and a combined approach that learned embeddings for characters based on their position in the word. This final representation yielded the largest improvements.

We use the same idea but augmented it with LSTM learned representations, and we enable interaction between the CRF and the LSTM parameters. More details are described in (§2.3).

## 2.3 Using Segmentation Representations to Improve NER

The improvements provided by character position embeddings demonstrated by Peng and Dredze (2015) indicated that word segmentation information can be helpful for NER. Embeddings aside, a simple way to include this information in an NER system would be to add features to the CRF using the predicted segmentation labels as features.

However, these features alone may overlook useful information from the segmentation model.

---

[2]The same functionality as $A_{ij}$ in the model of Chen et al. (2015).

Previous work showed that jointly learning different stages of the NLP pipeline helped for Chinese (Liu et al., 2012b; Zheng et al., 2013). We thus seek approaches for deeper interaction between word segmentation and NER models. The LSTM word segmentor learns two different types of representations: 1) embeddings for each character and 2) hidden vectors for predicting segmentation tags. Compressing these rich representations down to a small feature set imposes a bottleneck on using richer word segmentation related information for NER. We thus experiment with including both of these information sources directly into the NER model.

Since the log-bilinear CRF already supports joint training of lexical embeddings, we can also incorporate the LSTM output hidden vectors as dynamic features using a joint objective function.

First, we augment the CRF with the LSTM parameters as follows:

$$\mathcal{L}_n(\boldsymbol{y}_n; \boldsymbol{x}_n, \Theta) = \frac{1}{K} \sum_k \Big[ \log \frac{1}{Z(\boldsymbol{x}_n)^k} + \sum_j \Lambda_j F_j(\boldsymbol{y}_n^k, \boldsymbol{x}_n^k, \boldsymbol{e}_w, \boldsymbol{h}_w) \Big], \quad (3)$$

where $k$ indexes instances, $j$ positions, and

$$F_j(\boldsymbol{y}^k, \boldsymbol{x}^k, \boldsymbol{e_w}, \boldsymbol{h_w}) = \sum_{i=1}^n f_j(y_{i-1}^k, y_i^k, \boldsymbol{x}^k, \boldsymbol{e_w}, \boldsymbol{h_w}, i)$$

represents the feature functions. These features now depend on the embeddings learned by the LSTM ($\boldsymbol{e}_w$) and the LSTM's output hidden vectors ($\boldsymbol{h}_w$). Note that by including $\boldsymbol{h}_w$ alone we create dependence on all LSTM parameters on which the hidden states depend (i.e. the weight matrices). We experiment with including input embeddings and output hidden vectors independently, as well as both parameters together. An illustration of the integrated model is shown in Figure 1.

**Joint Training** In our integrated model, the LSTM parameters are used for both predicting word segmentations and NER. Therefore, we consider a joint training scheme. We maximize a (weighted) joint objective:

$$\mathcal{L}_{joint}(\Theta) = \lambda \mathcal{L}_s(\boldsymbol{y}_s; \boldsymbol{x}_s, \Theta) + \mathcal{L}_n(\boldsymbol{y}_n; \boldsymbol{x}_n, \Theta) \quad (4)$$

where $\lambda$ trades off between better segmentations or better NER, and $\Theta$ includes all parameters used in both models. Since we are interested in improving NER we consider settings with $\lambda < 1$.

|   | | Named Entity | | | | | Nominal Mention | | | | |
|---|---|---|---|---|---|---|---|---|---|---|---|
|   | | Dev | | | Test | | | Dev | | | Test | | |
|   | Method | Precision | Recall | F1 | Precision | Recall | F1 | Precision | Recall | F1 | Precision | Recall | F1 |
| 1 | CRF with baseline features | 60.27 | 25.43 | 35.77 | 57.47 | 25.77 | 35.59 | 72.06 | 32.56 | 44.85 | 59.84 | 23.55 | 33.80 |
| 2 | + Segment Features | 62.34 | 27.75 | 38.40 | 58.06 | 27.84 | 37.63 | 58.50 | 38.87 | 46.71 | 47.43 | 26.77 | 34.23 |
| 3 | P & D best NER model | 57.41 | 35.84 | 44.13 | 57.98 | 35.57 | 44.09 | 72.55 | 36.88 | 48.90 | 63.84 | 29.45 | 40.38 |
| 4 | + Segment Features | 47.40 | 42.20 | 44.65 | 48.08 | 38.66 | 42.86 | 76.38 | 36.54 | 49.44 | 63.36 | 26.77 | 37.64 |
| 5 | P & D w/ Char Embeddings | 58.76 | 32.95 | 42.22 | 57.89 | 34.02 | 42.86 | 66.88 | 35.55 | 46.42 | 55.15 | 29.35 | 38.32 |
| 6 | + Segment Features | 51.47 | 40.46 | 45.31 | 52.55 | 37.11 | 43.50 | 65.43 | 40.86 | 50.31 | 54.01 | 32.58 | 40.64 |
| 7 | Pipeline Seg. Repr. + NER | 64.71 | 38.14 | **48.00** | 64.22 | 36.08 | 46.20 | 69.36 | 39.87 | 50.63 | 56.52 | 33.55 | 42.11 |
| 8 | Jointly LSTM w/o feat. | 59.22 | 35.26 | 44.20 | 60.00 | 35.57 | 44.66 | 60.10 | 39.53 | 47.70 | 56.90 | 31.94 | 40.91 |
| 9 | Jointly Train Char. Emb. | 64.21 | 35.26 | 45.52 | 63.16 | 37.11 | 46.75 | 73.55 | 37.87 | 50.00 | 65.33 | 31.61 | 42.61 |
| 10 | Jointly Train LSTM Hidden | 61.86 | 34.68 | 44.44 | 63.03 | 38.66 | 47.92 | 67.23 | 39.53 | 49.79 | 60.00 | 33.87 | 43.30 |
| 11 | Jointly Train LSTM + Emb. | 59.29 | 38.73 | 46.85 | 63.33 | 39.18 | **48.41** | 61.61 | 43.19 | **50.78** | 58.59 | 37.42 | **45.67** |

Table 1: NER results for named and nominal mentions on dev and test data.

## 3 Parameter Estimation

We train all of our models using stochastic gradient descent (SGD.) We train for up to 30 epochs, stopping when NER results converged on dev data. We use a separate learning rate for each part of the joint objective, with a schedule that decays the learning rate by half if dev results do not improve after 5 consecutive epochs. Dropout is introduced in the input layer of LSTM following Chen et al. (2015). We optimize two hyper-parameters using held out dev data: the joint coefficient $\lambda$ in the interval $[0.5, 1]$ and the dropout rate in the interval $[0, 0.5]$. All other hyper-parameters were set to the values given by Chen et al. (2015) for the LSTM and Peng and Dredze (2015) for the CRF.

We train the joint model using an alternating optimization strategy. Since the segmentation dataset is significantly larger than the NER dataset, we subsample the former at each iteration to be the same size as the NER training data, with different subsamples in each iteration. We found subsampling critical and it significantly reduced training time and allowed us to better explore the hyper-parameter space.

We initialized LSTM input embeddings with pre-trained character-positional embeddings trained on 112,971,734 Weibo messages to initialize the input embeddings for LSTM. We used word2vec (Mikolov et al., 2013) with the same parameter settings as Peng and Dredze (2015) to pre-train the embeddings.

## 4 Experiments and Analysis

### 4.1 Datasets

We use the same training, development and test splits as Chen et al. (2015) for word segmentation and Peng and Dredze (2015) for NER.

**Word Segmentation** The segmentation data is taken from the SIGHAN 2005 shared task. We used the PKU portion, which includes 43,963 word sentences as training and 4,278 sentences as test. We did not apply any special preprocessing.

**NER** This dataset contains 1,890 Sina Weibo messages annotated with four entity types (person, organization, location and geo-political entity), including named and nominal mentions. We note that the word segmentation dataset is significantly larger than the NER data, which motivates our subsampling during training (§3).

### 4.2 Results and Analysis

Table 1 shows results for NER in terms of precision, recall and F1 for named (left) and nominal (right) mentions on both dev and test sets. The hyper-parameters are tuned on dev data and then applied on test. We now explain the results.

We begin by establishing a CRF baseline (#1) and show that adding segmentation features helps (#2). However, adding those features to the full model (with embeddings) in Peng and Dredze (2015) (#3) did not improve results (#4). This is probably because the character-positional embeddings already carry segmentation information. Replacing the character-positional embeddings with character embeddings (#5) gets worse results than (#3), but benefits from adding segmentation features (#6). This demonstrates both that word segmentation helps and that character-positional embeddings effectively convey word boundary information.

We now consider our model of jointly training the character embeddings (#9), the LSTM hidden vectors (#10) and both (#11). They all improve over the best published results (#3). Jointly training the LSTM hidden vectors (#10) does better than jointly training the embeddings (#9), probably because they carry richer word boundary information. Using both representations achieves the single best result (#11): 4.3% improvement on named and 5.3% on nominal mentions F1 scores.

Finally, we examine how much of the gain is from joint training versus from pre-trained segmentation representations. We first train an LSTM for word segmentation, then use the trained embeddings and hidden vectors as inputs to the log-bilinear CRF model for NER, and fine tune these representations. This (#7) improved test F1 by 2%, about half of the overall improvements from joint training.

## 5 Discussion

Huang et al. (2015) first proposed recurrent neural networks stacked with a CRF for sequential tagging tasks, as was applied to POS, chunking and NER tasks. More recent efforts have been made to add character level modeling and explore different types of RNNs (Lample et al., 2016; Ma and Hovy, 2016; Yang et al., 2016). These methods have achieved state-of-the-art results for NER on English news and several other Indo-European languages. However, this work has not considered languages that require word segmentation, nor do they consider social media.

We can view our method as multi-task learning (Caruana, 1997; Ando and Zhang, 2005; Collobert and Weston, 2008), where we are using the same learned representations (embeddings and hidden vectors) for two tasks: segmentation and NER, which use different prediction and decoding layers. Result #8 shows the effect of excluding the additional NER features and just sharing a jointly trained LSTM[3]. While this does not perform as well as adding the additional NER features (#11), it is impressive that this simple architecture achieved similar F1 as the best results in Peng and Dredze (2015). While we may expect both NER and word segmentation results to improve, we found the segmentation performances of the best joint model tuned for NER lose to the stand alone word segmentation model (F1 of 90.7% v.s. 93.3%). This lies in the fact that tuning $\lambda$ means choosing between the two tasks; no single setting achieved improvements for both, which suggests further work is needed on better model structures

---

[3]This reduces to the multi-task setting of Yang et al. (2016).

and learning.

Second, our segmentation data is from the news domain, whereas the NER data is from social media. While it is well known that segmentation systems trained on news do worse on social media (Duan et al., 2012), we still show large improvements in applying our model to these different domains. It may be that we are able to obtain better results in the case of domain mismatch because we integrate the representations of the LSTM model directly into our CRF, as opposed to only using the predictions of the LSTM segmentation model. We plan to consider expanding our model to explicitly include domain adaptation mechanisms (Yang and Eisenstein, 2015).